# Multi-frame image super-resolution with fast upscaling technique

Longguang Wang, Zaiping Lin, Xinpu Deng, Wei An[1]

*Abstract*—Multi-frame image super-resolution (MISR) aims to fuse information in low-resolution (LR) image sequence to compose a high-resolution (HR) one, which is applied extensively in many areas recently. Different with single image super-resolution (SISR), sub-pixel transitions between multiple frames introduce additional information, attaching more significance to fusion operator to alleviate the ill-posedness of MISR. For reconstruction-based approaches, the inevitable projection of reconstruction errors from LR space to HR space is commonly tackled by an interpolation operator, however crude interpolation may not fit the natural image and generate annoying blurring artifacts, especially after fusion operator. In this paper, we propose an end-to-end fast upscaling technique to replace the interpolation operator, design upscaling filters in LR space for periodic sub-locations respectively and shuffle the filter results to derive the final reconstruction errors in HR space. The proposed fast upscaling technique not only reduce the computational complexity of the upscaling operation by utilizing shuffling operation to avoid complex operation in HR space, but also realize superior performance with fewer blurring artifacts. Extensive experimental results demonstrate the effectiveness and efficiency of the proposed technique, whilst, combining the proposed technique with bilateral total variation (BTV) regularization, the MISR approach outperforms state-of-the-art methods.

*Index Terms*—multi-frame super-resolution, upscaling technique, bilateral total variation, shuffling operation

## I. INTRODUCTION

DUE to the limited technical and manufacturing level, the resolution of image may not be satisfied in video surveillance [1][2], medical imaging [3][4][5], aerospace [6][7][8] and many other fields [9][10], where high resolution (HR) images are commonly required and desired for distinct image details. With charge-coupled device (CCD) and CMOS image sensors developing rapidly in recent decades, the increasing demand in image resolution still cannot be satisfied, leading to attempts to steer clear of the sensor issues but utilize computational imaging to improve the spatial resolution, namely super-resolution (SR).

Serving as a typical inverse problem, SR aims to recover missing image details during image degradation process, which is underdetermined, requiring additional information to alleviate the ill-posedness. For single image super-resolution (SISR), lack of observation information leads to attempts to exploit additional information to learn how natural images are, and many example-based approaches [11][12][13][14][15][16] have been proposed. For multi-frame image super-resolution (MISR), sub-pixel transitions between multiple observations provide underlying available information, therefore reconstruction-based approach [17][20][21][25][26] is mainly concentrated on to derive the high resolution (HR) image through maintaining global consistency, which is intuitive and natural.

Concerning reconstruction-based MISR approaches, extensive works have been put forward focusing on the design of regularization to realize favorable results. Tikhonov regularization SR method [17][18], as a representative method, introduces smoothness constraints to suppress the noise but results in the loss of detailed information in edge regions. To realize edge preserving, total variation (TV) operator is introduced as a regularization term [19][20][21], however, leads to the deterioration of smoothness in local flat region. Motivated by bilateral filter, Farsiu et al. [25] proposed the bilateral total variation (BTV) operator measured by $L_1$ norm, which integrates TV with bilateral filter and realizes superior performance and robustness. Due to the performance and simplicity of BTV, further improvement has attracted extensive investigation, Li et al. [31] proposed the locally adaptive bilateral total variation (LABTV) operator measured by fuzzy-entropy-based neighborhood homogeneous measurement, realizing locally adaptive regularization.

Among these reconstruction-based approaches, to maintain global consistency with multiple observations, reconstruction errors are commonly integrated in the cost function to penalize the discrepancy between reconstructed HR image and LR observations. Within the iterative SR process, inevitable projection of reconstruction error from LR space to HR space is usually tackled by an interpolation operator for simplicity, however this crude operation may introduce additional errors and lead to deteriorated convergence and performance, especially after fusion operation of MISR. In this paper, we propose an end-to-end fast upscaling technique to replace the interpolation operation in the SR framework. Firstly, we unfold the degradation model to analyze underlying contributions of periodic sub-locations to reconstruction error in LR space and design upscaling filters correspondingly. Secondly, the filter results utilizing designed upscaling filters are shuffled to derive the reconstruction errors in HR space. Finally, the reconstruction errors are cooperated with regularization term to modify the HR image iteratively until convergence. Extensive experiments are conducted to demonstrate the effectiveness and efficiency of the proposed upscaling technique, besides, combining the proposed technique with BTV regularization, the MISR approach realizes state-of-the-art performance.

The rest of the paper is organized as follows: **Section II** mainly formulates the problem of multi-frame image super-resolution. **Section III** presents the proposed upscaling technique in detail. **Section IV** performs extensive experiments compared with other state-of-the-art approaches and the conclusions are drawn in **Section V**.

[1] L. Wang is with the College of Electronic Science and Engineering, National University of Defense Technology, Changsha, China. (e-mail: wanglongguang15@nudt.edu..cn)

Z. Lin, X. Deng, and W. An are also with the College of Electronic Science and Engineering, National University of Defense Technology, Changsha, China.

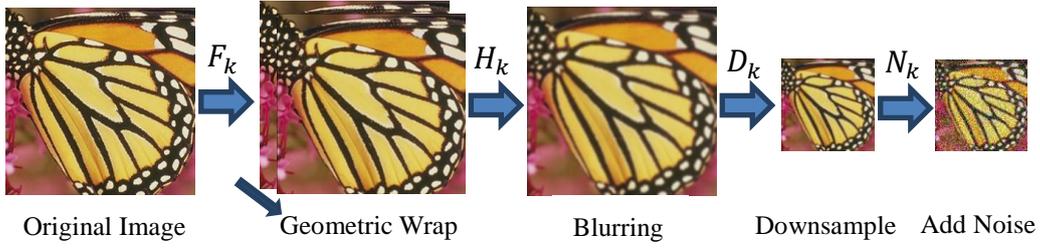

Fig. 1. Sketch of degradation model.

## II. MULTI-FRAME IMAGE SR PROBLEM FORMULATION

### A. Degradation model

As the inverse process of image degradation, SR reconstruction is tightly dependent on the degradation model. With many degrading factors existing like atmospheric turbulence, optical blurring, relative motion and sampling process, the degradation model of LR images can be formulated as:

$$Y_k = D_k H_k F_k X + N_k, k = 1,2,...M, \quad (1)$$

where $Y_k$, $X$ represent $k^{th}$ LR image and HR image respectively, $D_k, H_k, F_k$ serve as decimation matrix, blurring operator and geometric warp matrix of $Y_k$ respectively, and $N_k$ is the additional Gaussian noise in $Y_k$. Note that although complex motions may be common in real sequences which cannot be represented by a simple parametric form and many works [28][29][30] tend to address this problem, global translational displacements between multiple frames serving as a fundamental issue, is still the focus of this paper.

Generally assuming all LR images are generated under the same condition, we can derive the following model:

$$Y_k = DHF_k X + N_k, k = 1,2,...M, \quad (2)$$

where $D$ and $H$ represent same decimation matrix and blurring operator respectively in all LR images. The degradation model is further illustrated in **Fig. 1**.

### B. Reconstruction-based SR process

In Bayesian framework, SR reconstruction is equivalent to the estimation of HR image with given LR images, where maximum a posteriori (MAP) estimator is extensively utilized as

$$X = argmax_X \{\prod_{k=1}^M p(X|Y_k)\}$$
$$= argmax_X \{\prod_{k=1}^M p(Y_k|X) \cdot p(X)\}. \quad (3)$$

To solve the probabilistic maximization problem, equivalent minimization of reconstruction errors can be derived as:

$$X = argmin_X \{\sum_{k=1}^M \|DHF_k X - Y_k\|^2\}. \quad (4)$$

As insufficient information given in LR image sequence $Y_k$, reconstructing the original HR image $X$ is an underdetermined problem. To solve the ill-posed problem, regularization is commonly introduced as priori knowledge to obtain a stable solution and (4) can be rewritten as:

$$X = argmin_X \{\sum_{k=1}^M \|DHF_k X - Y_k\|^2 + \lambda R(X)\}, \quad (5)$$

where $R(X)$ is the regularization term of HR image $X$, $\lambda$ serves as a regularization parameter weighting the reconstruction errors against the regularization cost.

Assuming decimation matrix $D$, blurring operator $H$ and geometric warp matrix $F_k$ are already known, the minimization problem can be solved utilizing steepest descent approach as:

$$X^{l+1} = X^l - \eta \left(\sum_{k=1}^M F_k^T H^T D^T (DHF_k X - Y_k) + \lambda \frac{\partial R(X)}{\partial X}\right)$$
(6)

where $X^l$, $X^{l+1}$ are estimators of HR image $X$ in $l^{th}$ and $l + 1^{th}$ iteration respectively, $\eta$ is the learning rate representing the pace to approach the optimal during iterations. In this paper, the derivation of displacements and blur kernel is not under consideration, we assume the blur kernel is already known and utilize optical flow method to estimate the underlying displacements.

## III. MULTI-FRAME IMAGE SR WITH FAST UPSCALING TECHNIQUE

In this section, we first present the proposed end-to-end fast upscaling technique, introduce our motivation and its formulation in detail before theoretical analysis on computational complexity and convergence. Then we integrate the proposed upscaling technique with BTV regularization to construct the overall MISR framework.

### A. End-to-end upscaling technique

#### 1) Motivation

As we can see from (6), projecting reconstruction errors $(DHF_k X - Y_k)$ from LR space into HR space is required in the inference, where interpolation operator commonly plays a main role as the upscaling operator $D^T$, and then deblurring operator $H^T$ and inverse translation operator $F_k^T$ are performed in HR space. Lacking in theoretical basis, crude interpolation may introduce additional errors leading to blurring artifacts, therefore it serves as a fundamental operator and requires small stepsize and enough iterations to alleviate the deterioration, which is sub-optimal and adds computational complexity [27].

In Shi et al. [27], an efficient sub-pixel convolutional neural network is proposed, where an array of upscaling filters are utilized to corporate with a shuffling operator to upscale the final LR feature maps in to HR output, which is located at the very end of the network. As [27] demonstrates increasing the resolution of LR image before image enhancement increase the computational complexity, besides the commonly used interpolation methods do not bring additional information to solve the ill-posed reconstruction problem, we are inspired to unfold the degradation model to analyze underlying contributions of periodic sub-locations to reconstruction error in LR space and design similar array of upscaling filters. In this paper, we propose an end-to-end upscaling technique to perform fast and efficient upscaling operation, serving as a direct bridge between reconstruction errors in LR space and HR space.

#### 2) Formulation

Further analyze the degradation model shown in (2) from the perspective of image, assuming the blurring operator $H$ is limited in a $a \times a$ region ($a$ is odd for the symmetry of blurring kernel) in HR space and the upscaling factor is determined as $\gamma$, namely the decimation operator $\boldsymbol{D}$ is limited in a $\gamma \times \gamma$ region in HR space. Concerning translational displacements $\boldsymbol{d}$ between LR images, only sub-pixel displacements $\boldsymbol{d}_{subpixel}$ are considered for pixel-level displacements do not bring additional information. Without loss of generality, only positive sub-pixel displacements $\boldsymbol{d}_{subpixel}$ ($\boldsymbol{d}_{subpixel} = \boldsymbol{d} - floor(\boldsymbol{d})$) are taken into consideration, so that the geometric wrap operator can be limited in a $2 \times 2$ region.

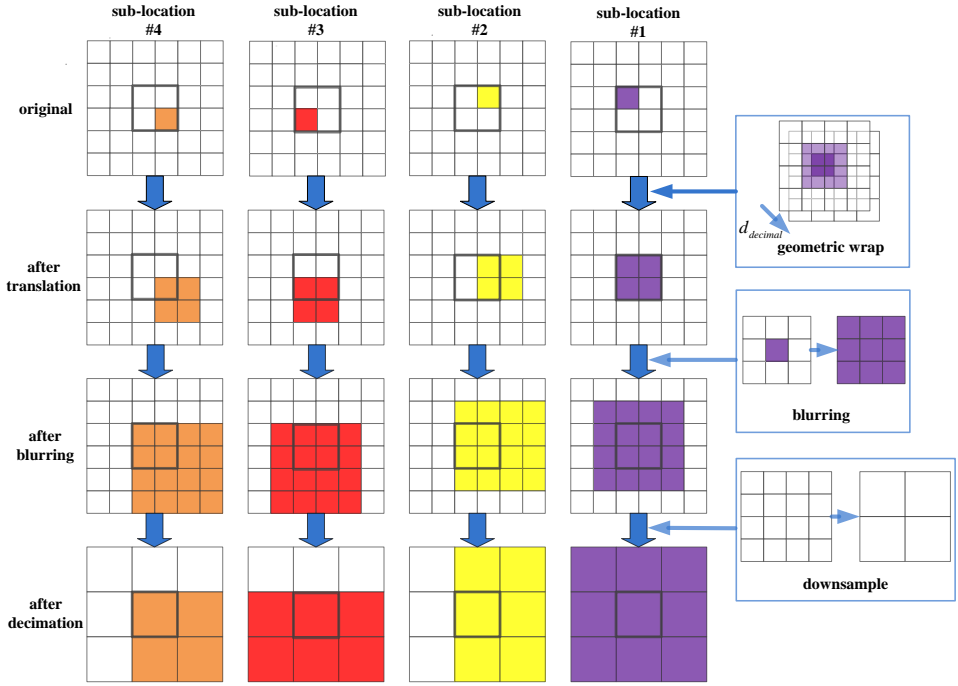

Fig. 2. Degradation process with respect to different sub-locations.

To illustrate the interaction between the concatenation of operators, here we set $a$ to be 3 with $\gamma$ to be 2 and derive the degradation model from the perspective of image as shown in **Fig. 2**.

With the degradation process unfolded as shown in **Fig. 2**, it can been seen that different ranges of influence in LR space correspond with different sub-locations, which inspires us to structure upscaling filters concerning periodic sub-locations utilizing the differences between influence ranges. As parameters including sub-pixel displacements $\boldsymbol{d}_{subpixel}$, blurring kernel $\boldsymbol{H}$ and upscaling factor $\gamma$ are all determined, the overall degradation process and the underlying contributions of periodic sub-locations to LR space can be derived.

Remembering the projection of reconstruction errors from LR space to HR space in (6), we structure upscaling filters utilizing the underlying contributions of periodic sub-locations to realize end-to-end upscaling of reconstruction errors in LR space. Within the probabilistic framework, the upscaling operator can be equivalent to an optimal estimation problem as

$$g_i = \underset{h_i}{argmax}\, p(h_i|e_{j,j\in C(i)}), i = 1,2,\dots,N_{HR}, \quad (7)$$

where $g_i$, $e_j$ are reconstruction error of $i^{th}$ pixel in HR space and $j^{th}$ pixel in LR space respectively, $C(i)$ serves as the influence range in LR space of pixel $i$ in HR space and $N_{HR}$ ($N_{HR} = \gamma^2 N_{LR}$) is the number of pixels in HR space. Further assuming $p_{ij}$ serves as the contribution of $i^{th}$ pixel in HR space to $j^{th}$ pixel in LR space, minimization problem equivalent to (7) can be derived utilizing greedy strategy

$$g_i = \underset{h_i}{argmin}\{\sum_{j=1}^{N_{C(i)}}\|p_{ij}h_i - e_j\|_p^p\}, i = 1,2,\dots,N_{HR}, \quad (8)$$

where $N_{C(i)}$ is the number of pixels within $C(i)$. As the influence ranges of different sub-locations can be limited in a $3\times 3$ LR region in the case of **Fig. 2**, here we utilize $L_2$ norm for simplicity and the closed-form solution can be computed as

$$h_i = \frac{\sum_{j=1}^{9} p_{ij} e_j}{\sum_{j=1}^{9} p_{ij}^2}, i = 1,2,\dots,N_{HR}. \quad (9)$$

Considering influence range and corresponding contributions are both sub-location dependent, namely HR pixels with same sub-location share identical influence range and contribution distribution, we intuitively separate the upscaling operator with respect to different sub-locations and rewrite (9) for identical sub-location in a convolution form due to the global consistent process.

$$\boldsymbol{G}_i = \frac{T_i * \boldsymbol{E}}{\|T_i\|_2^2} = T'_i * \boldsymbol{E}, i = 1,2,\dots,N_{HR}, \quad (10)$$

where $\boldsymbol{G}_i$ represents reconstruction error map for $i^{th}$ sub-location in HR space, $\boldsymbol{E}$ represents reconstruction error map in LR space, namely $\boldsymbol{E} = \boldsymbol{D}\boldsymbol{H}\boldsymbol{F}_k\boldsymbol{X} - \boldsymbol{Y}_k$ and $T_i$ serves as the contribution distribution concerning $i^{th}$ sub-location. Regarding the $L_2$ norm of $T_i$ as a normalization constant, we integrate it into $T_i$ to derive normalized contribution distribution $T'_i$ as filter masks. In this way, the upscaling operator can be implemented by convolution operator, which realizes favorable efficiency.

As reconstruction errors with respect to ranged sub-locations in HR space derived separately, a shuffling operator is introduced to rearrange the elements in separate error maps to a complete error map $\boldsymbol{G}$ in HR space as shown in **Fig. 3**. Utilizing the proposed end-to-end upscaling technique, we evade the interpolation operator which may introduce additional errors, design filter masks according to contribution distribution concerning ranged sub-locations and process the error map in LR space separately, finally a shuffling operator is implemented to derive the final error map in HR space. As all processes in LR space are end-to-end, namely all processing results can be mapped directly to corresponding HR space without intermediate operations, our upscaling technique can realize superior efficiency and effectiveness, which is demonstrated in the following analysis and Section IV.

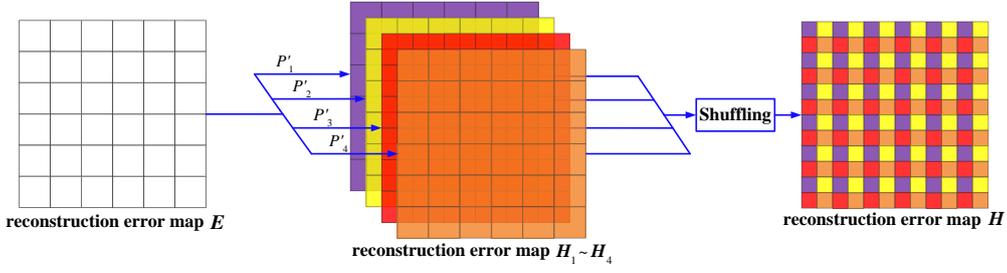

Fig. 3. End-to-end upscaling technique.

*3) Theoretical analysis*

In this section, theoretical analysis with respect to computational complexity and convergence are carried out respectively, we attempt to illustrate the superiority of the proposed end-to-end upscaling technique theoretically.

● **Computational complexity**

For conventional interpolation-based upscaling technique, the reconstruction errors in LR space are commonly projected into HR space by an interpolation operator first, and then processed by deblurring operator $H^T$ and inverse translation operator $F_k^T$. In this way, deblurring operator and inverse translation operator are both performed in HR space, which adds computational complexity.

Although the complexities of interpolation-based upscaling technique and the proposed end-to-end upscaling technique are both of order $O(N_{HR})$, where $N_{HR}$ is the number of pixels in HR space, the computation amounts differ greatly. Assuming $H, F_k$ are limited in $a \times a$ and $2 \times 2$ region of HR space respectively, for interpolation-based upscaling technique, bicubic interpolation is commonly utilized performing as weighted sum of neighboring $4 \times 4$ pixels in LR space, afterwards deblurring operator $H^T$ and inverse translation operator $F_k^T$ are performed as weighted sum of neighboring $a \times a$ and $2 \times 2$ pixels respectively in HR space. For our proposed end-to-end upscaling technique, the upscaling operator is performed as weighted sum of neighboring $ceil(\frac{a+1}{\gamma} + 1) \times ceil(\frac{a+1}{\gamma} + 1)$ pixels in LR space, which remarkably reduces the computational complexity especially with larger upscaling factor $\gamma$.

● **Convergence**

For reconstruction-based MISR approaches, steepest descent and fixed stepsize are commonly utilized for interpolation-based upscaling technique, which requires small stepsize and enough iterations to approach the optimal. As interpolation-based upscaling technique introduces additional errors, the deviation of descent direction makes the convergence process greatly time consuming.

While Newton-type methods typically tend to converge in fewer iterations, the computation of Hessian matrix in each iteration is required, leading to expensive computational cost. As we analyze our end-to-end upscaling technique theoretically, it can be regarded as a variation and simplification of Newton-type method, which can realize superior convergence.

Remembering the minimization problem in (5), as we unfold the degradation model, it can be rewritten as

$$X = argmin_X\{\sum_{k=1}^{N}\|A_k X - Y_k\|^2 + \lambda R(X)\}, \quad (11)$$

where $X, Y_k$ represent vectorized HR image and $k^{th}$ LR image respectively, $A_k = [a_{k1}, a_{k2}, ..., a_{kL}]$ serves as a dictionary arranged in lexicographic order, which consists of $N_{HR}$ atoms. As analyzed before, different sub-locations correspond to different influence ranges and contribution distributions, we utilize this characteristic to construct overcomplete dictionary as shown in **Fig. 4**.

For Newton method, the inference of $X$ can be written as

$$X^{l+1} = X^l - \sum_{k=1}^{M} \frac{2A_k^T(A_kX-Y_k)+\lambda\frac{\partial R(X)}{\partial X}}{2A_k^T A_k + \lambda\frac{\partial^2 R(X)}{\partial X^2}}. \quad (12)$$

As dictionary $A_k$ is hard to manipulate, Newton method cannot be directly utilized in general. Considering $R(X)$ is commonly a non-linear operator and the computation of second derivation is relatively difficult, besides the regularization parameter $\lambda$ is usually small, we simplify (12) and separate $R(X)$ term out as

$$X^{l+1} = X^l - \left(\sum_{k=1}^{M} \frac{A_k^T(A_kX-Y_k)}{A_k^T A_k} + \lambda \frac{\frac{\partial R(X)}{\partial X}}{2A_k^T A_k}\right). \quad (13)$$

Unfold the symmetric matrix $A_k^T A_k$ and we can derive

$$\begin{bmatrix} a_{k1}^T a_{k1} & a_{k1}^T a_{k2} & \cdots & a_{k1}^T a_{k-1} & a_{k1}^T a_{kN_{HR}} \\ a_{k2}^T a_{k1} & a_{k2}^T a_{k2} & & a_{k2}^T a_{K-1} & a_{k2}^T a_{kN_{HR}} \\ \vdots & & \ddots & & \vdots \\ a_{k(N_{HR}-1)}^T a_{k1} & a_{k(N_{HR}-1)}^T a_{k2} & \cdots & a_{k(N_{HR}-1)}^T a_{k(N_{HR}-1)} & a_{k(N_{HR}-1)}^T a_{kL} \\ a_{kN_{HR}}^T a_{k1} & a_{kN_{HR}}^T a_{k2} & & a_{kN_{HR}}^T a_{k(N_{HR}-1)} & a_{kN_{HR}}^T a_{kN_{HR}} \end{bmatrix}. \quad (14)$$

As we can see from **Fig. 4**, the atom is highly sparse and regional, namely $a_{ki}^T a_{kj}$ equals to zero except $j \in \eta(i)$, where $\eta(i)$ represents the neighborhood of corresponding HR pixel $i$ in HR space. Taking this into consideration, elements in $A^T A$ can be rearranged through placing correlative atoms closer, and then we can derive an approximate diagonal matrix, namely most entries in $A_k^T A_k$ equal to zeros except diagonal ones and some other ones.

As the inverse operation of $A_k^T A_k$ is hard to manipulate, we only take diagonal elements into consideration, namely regard $A_k^T A_k$ as a diagonal matrix by ignoring other non-diagonal non-zero entries, and rewrite (13) as

$$X^{l+1} = X^l - \left(\sum_{k=1}^{M} \frac{A_k^T(A_kX-Y_k)}{diag(A_k^T A_k)} + \lambda \frac{\frac{\partial R(X)}{\partial X}}{2 diag(A_k^T A_k)}\right). \quad (15)$$

Rewrite $\frac{A_k^T(A_kX-Y_k)}{diag(A_k^T A_k)}$ in an atom-wise way and we can

derive

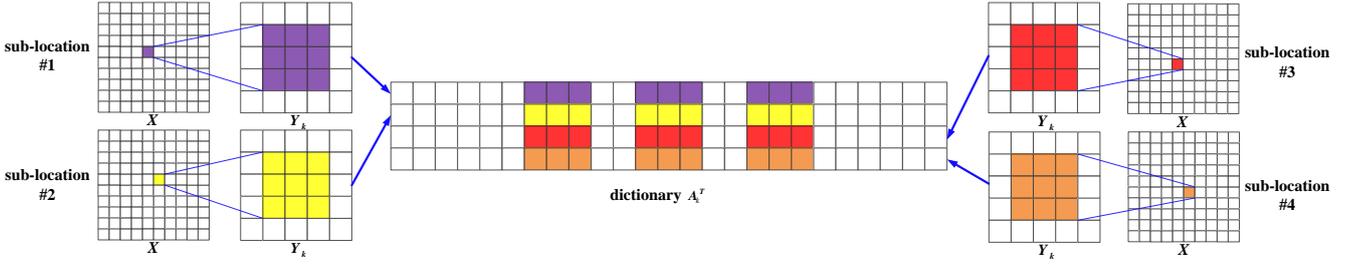

Fig. 4. Procedure of dictionary $A_k$.

$$\left(\frac{A_k^T(A_kX-Y_k)}{diag(A_k^TA_k)}\right)_i = \frac{a_{ki}^T(A_kX-Y_k)}{a_{ki}^Ta_{ki}}. \quad (16)$$

If we push atom $a_{ki}$ backwards into the corresponding LR image, (16) can rewritten in a convolution form

$$\left(\frac{A_k^T(A_kX-Y_k)}{diag(A_k^TA_k)}\right)_i = \frac{T_{ki}*(A_kX-Y_k)}{\|T_{ki}\|_2^2}, \quad (17)$$

where $T_{ki}$ represents the contribution distribution map corresponding to atom $a_{ki}$. Now we can see (15) performs identical to (10) concerning reconstruction error in LR space, illustrating the proposed end-to-end upscaling technique performs as a variation and simplification of Newton method, which can realize superior convergence.

As $a_{ki}$ is a one-sum atom, $\frac{1}{\|T_{ki}\|_2^2}$ performs as a magnification constant no less than 1. Considering $\lambda$ is commonly small, the magnification effect on regularization term can be ignored and we can derive

$$X^{l+1} = X^l - \left(\sum_{k=1}^M \frac{A_k^T(A_kX-Y_k)}{A_k^TA_k} + \lambda\frac{\partial R(X)}{\partial X}\right). \quad (18)$$

Considering our end-to-end upscaling technique performs as an approximate Newton method where the simplifications may introduce additional errors, therefore we also applied a similar learning rate $\eta$ in (18) as (6) for stable convergence

$$X^{l+1} = X^l - \eta\left(\sum_{k=1}^M \frac{A_k^T(A_kX-Y_k)}{diag(A_k^TA_k)} + \lambda\frac{\partial R(X)}{\partial X}\right). \quad (19)$$

However, the descent direction utilizing our end-to-end upscaling technique is relatively more accurate, therefore the convergence can be faster and more stable.

To further demonstrate the superior convergence, we utilize Tikhonov regularization without loss of generality and compare the convergence process of interpolation-based with the end-to-end upscaling technique, the comparison of convergence process is shown in **Fig. 5**.

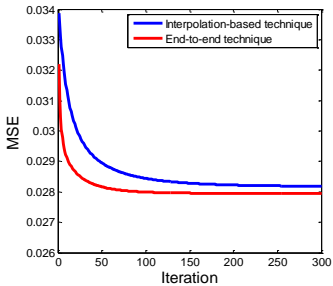

Fig. 5. Comparison of convergence process

As we can see form **Fig. 5**, our technique converges within around 50 iterations while interpolation-based technique requires more than 150 iterations to converge, demonstrating superior convergence of our end-to-end upscaling technique.

B. *MISR framework*

In this section, we integrate the proposed end-to-end upscaling technique with BTV regularization to construct the overall MISR framework. Due to the performance and simplicity of BTV, it has become one of the most commonly applied regularization in SR process, therefore we utilize BTV in our MISR framework to corporate with the proposed end-to-end upscaling technique. The overall framework is illustrated in **Fig. 6** and further summarized in **Algorithm 1**.

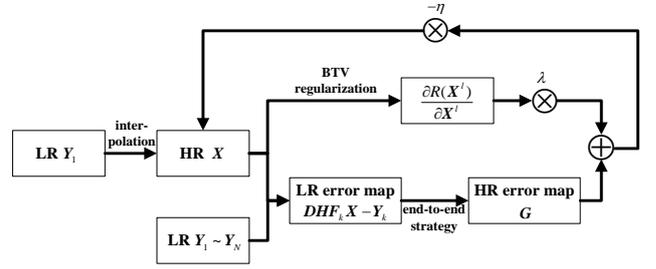

Fig. 6. Overall MISR framework

*Algorithm 1: MISR utilizing end-to-end upscaling technique*

*Input:* LR images $Y_1, Y_2, ..., Y_M$, blurring kernel, upscaling factor $\gamma$
*Initialize:* Select target image ($Y_1$ for example), utilize bicubic interpolation to derive initial HR image $X^0$, estimate translational displacements $d_{1k}$ between target image $Y_1$ and other LR images.
*Loop until* $\|X^l - X^{l-1}\|_2 / \|X^{l-1}\|_2 < \epsilon$ or $l \geq N_{Max}$
  ➢ Compute error map $DHF_kX - Y_k$ in LR space respectively
  ➢ Perform end-to-end upscaling technique to derive error map $G$ in HR space
  ➢ Compute BTV regularization $R(X)$ and its derivation $\frac{\partial R(X)}{\partial X}$
  ➢ Update $X^{l-1}$ to derive $X^l$ according to (19)
*Output:* Reconstructed HR image $X$

IV. EXPERIMENTAL RESULTS

In this section, extensive experimental results are presented to demonstrate the effectiveness and efficiency of the proposed end-to-end upscaling technique. We first perform experiments to demonstrate the effectiveness of our upscaling technique through equipping it to various reconstruction-based MISR methods, and then the proposed MISR framework is compared with other state-of-the-art algorithms.

As described in the degradation model, the degraded LR images are generated from an HR image through parallel translations, blurring, downsampling and addition of noises. In the experiments, the translational displacements $d$ are randomly set with vertical and horizontal shifts randomly sampled from a uniform distribution in [-5, 5]. The blurring operator is realized utilizing a $5 \times 5$ Gaussian kernel with standard deviation $\sigma = 1.2$. After geometric wrapping and blurring operation, the images are then downsampled by a factor $\gamma = 3$. Finally Gaussian noise with standard deviation $\sigma_{noise} = 1$ is added.

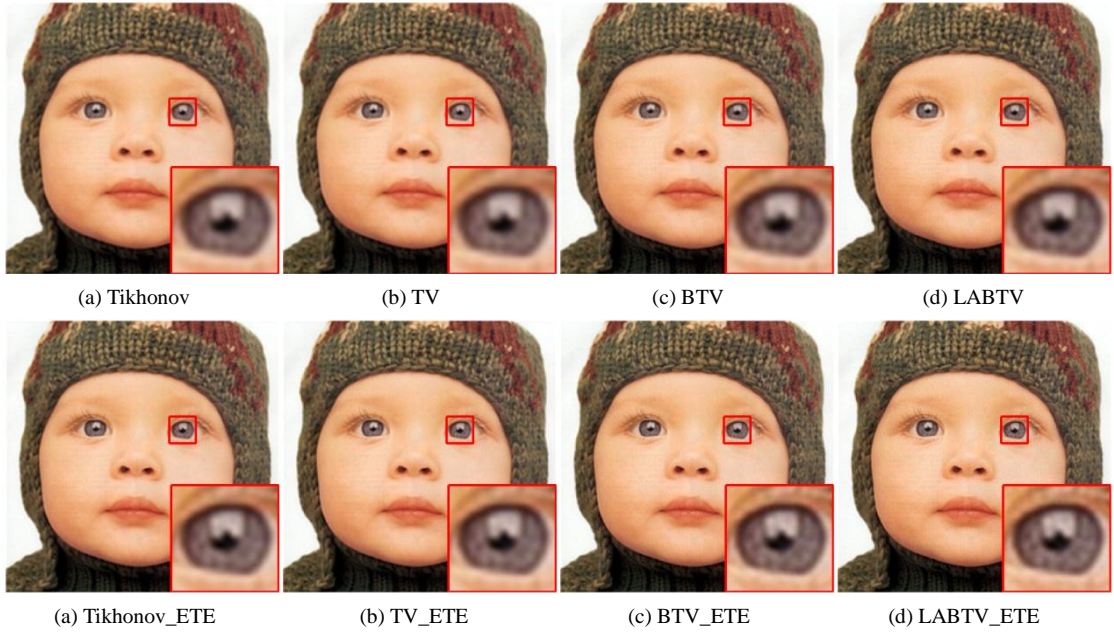

Fig. 7. Comparison of methods utilizing end-to-end technique with baselines on image *Baby*.

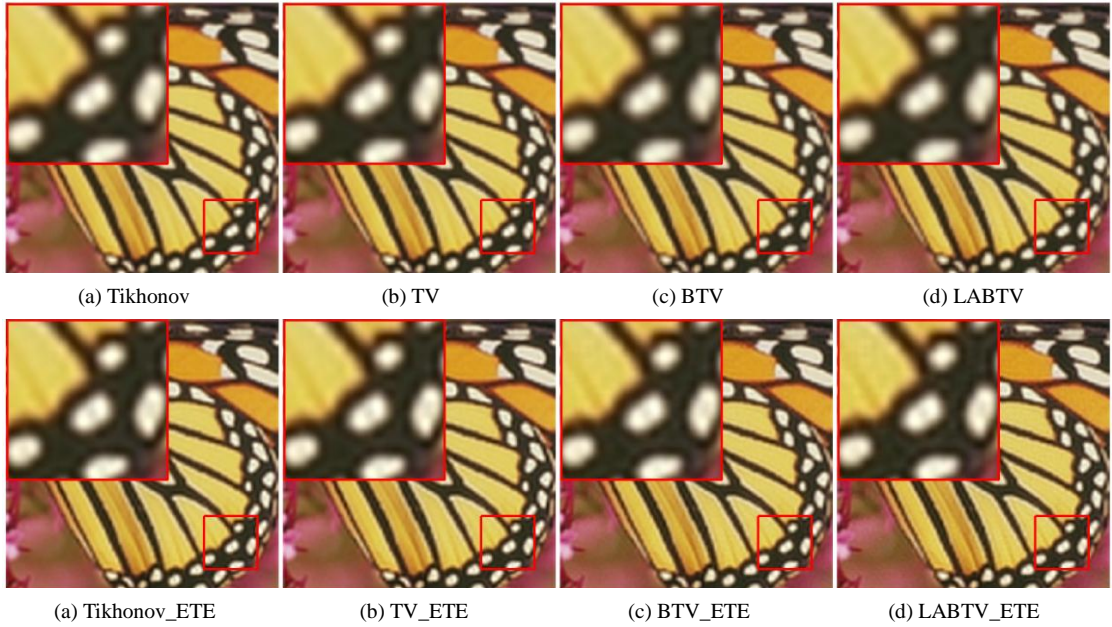

Fig. 8. Comparison of methods utilizing end-to-end technique with baselines on image *Butterfly*.

In our scenario, we use 5 LR images to reconstruct an HR image for MISR approaches, select the first one as target image without loss of generality and suppose the blurring kernel is already given. As human vision is more sensitive to brightness changes, all the SR methods are implemented only in brightness channel (Y) with color channels (UV) upscaled by bicubic interpolation for color images. All the experiments are coded in Matlab R2011 and running on a workstation with Septuple Core i7 @ 3.60 GHz CPUs and 16 GB memory.

For quantitative analysis and comparison of reconstruction performance, peak-signal-to-noise ratio (PSNR) and mean structure similarity (SSIM) are utilized as metrics, which are defined as

$$\begin{cases} PSNR = 10log_{10}\left(\frac{255^2}{MSE}\right), MSE = \frac{1}{MN}\sum_{i=1}^{M}\sum_{j=1}^{N}(Y(i,j) - X(i,j))^2 \\ SSIM = \frac{(2\mu_x\mu_y+c_1)(2\sigma_x\sigma_y+c_2)}{(\mu_x^2+\mu_y^2+c_1)(\sigma_x^2+\sigma_y^2+c_2)}, \begin{cases} c_1 = (k_1L)^2 \\ c_2 = (k_2L)^2 \end{cases} \end{cases}, \quad (20)$$

where $\mu_x, \mu_y$ are mean value of image $X$ and $Y$ respectively, $\sigma_x, \sigma_y$ are standard variance of $X$ and $Y$ respectively, $c_1, c_2$ are two stabilizing constants with $L$ representing the dynamics of a pixel value, and $k_1, k_2$ are generally set to be 0.01 and 0.03 respectively.

## A. Evaluation of the proposed upscaling technique

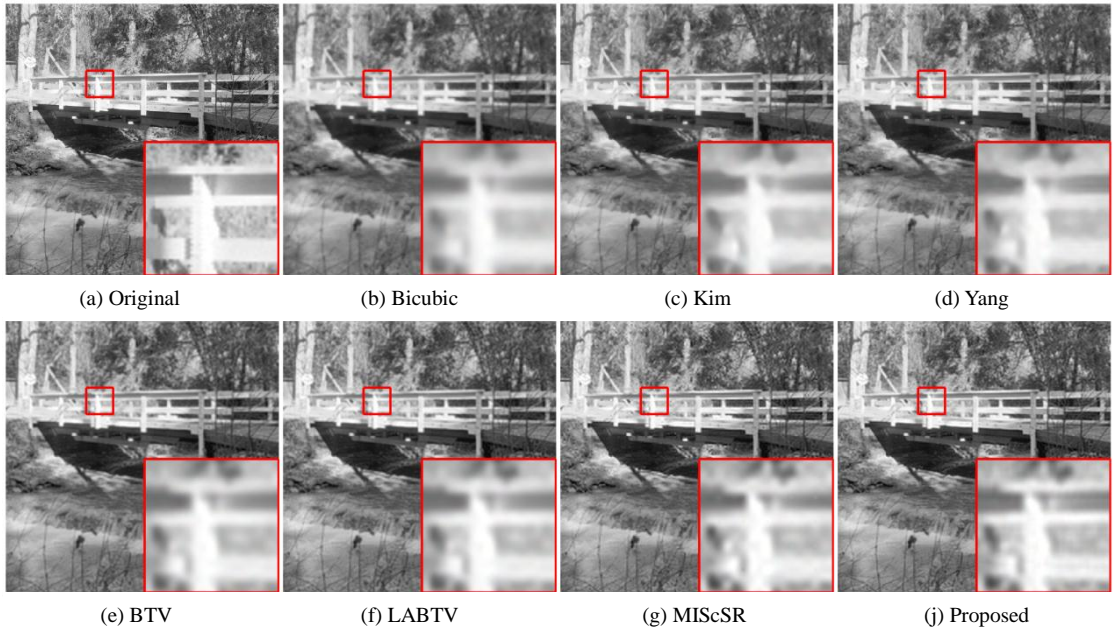

Fig. 9. Reconstruction results for image *Bridge* by ranged methods.

To validate the effectiveness and efficiency of the proposed end-to-end upscaling technique, we first select four representative reconstruction-based MISR method consisting of Tikhonov, TV [19], BTV [22] and LABTV [24] method as baseline methods, apply our technique to replace interpolation-based technique in the SR pipelines (note as Tikhonov_ETE for example) and conduct experiments on Set5 dataset to compare the performance correspondingly. Visual comparison is exhibited in **Fig. 7** and **8** with quantitative results are presented in **Table. I**.

As we can see from **Fig. 7** and **8**, compared with corresponding baseline methods, methods equipped with our end-to-end technique generate sharper edges and fine details, effectively alleviate the blurring effects with fewer artifacts and realize superior visual quality, which demonstrates the effectiveness of our end-to-end technique.

From the quantitative results shown in **Table. I** we can further see that the proposed technique remarkably improves the reconstruction performance with respect to PSNR and SSIM in all the images of Set5, whilst accelerating the MISR process. The PSNR values have been improved by around 0.9 (dB) in average while SSIM values also increased by around 0.015. Concerning computational complexity, the running time of equipped methods is shortened by around 0.3s in average with practicability greatly enhanced.

TABLE I
COMPARISON OF PSNR, SSIM AND RUNNDING TIME. THE MEAN PERFROMANCE OF 50 EXPERIMENTS IS PRESENTED WITH THE PERFORMANCE IMPROVEMENT EQUIPPED WITH THE PROPOSED TECHNIQUE SHOWN IN BRACKETS (RED AND BOLD).

| Set5 | Metric | Tikhonov | Tikhonov_ETE | TV | TV_ETE | BTV | BTV_ETE | LABTV | LABTV_ETE |
|---|---|---|---|---|---|---|---|---|---|
| baby | PSNR | 32.75 | 33.60(↑**0.85**) | 32.76 | 33.61(↑**0.85**) | 33.18 | 33.91(↑**0.73**) | 33.38 | 33.99(↑**0.61**) |
|  | SSIM | 0.887 | 0.903(↑**0.016**) | 0.886 | 0.903(↑**0.017**) | 0.898 | 0.903(↑**0.005**) | 0.901 | 0.904(↑**0.003**) |
|  | Time | 1.67 | 1.19(↓**0.48**) | 2.03 | 1.50(↓**0.53**) | 2.87 | 2.34(↓**0.53**) | 25.92 | 25.39(↓**0.53**) |
| bird | PSNR | 30.64 | 31.59(↑**0.95**) | 30.66 | 31.61(↑**0.95**) | 30.80 | 31.99(↑**1.19**) | 31.00 | 31.99(↑**0.99**) |
|  | SSIM | 0.898 | 0.909(↑**0.011**) | 0.898 | 0.909(↑**0.011**) | 0.903 | 0.911(↑**0.002**) | 0.904 | 0.906(↑**0.002**) |
|  | Time | 0.75 | 0.47(↓**0.28**) | 0.80 | 0.58(↓**0.22**) | 1.06 | 0.82(↓**0.24**) | 8.70 | 8.47(↓**0.23**) |
| butterfly | PSNR | 23.15 | 24.12(↑**0.97**) | 23.19 | 24.13(↑**0.94**) | 22.78 | 24.08(↑**1.30**) | 23.01 | 24.23(↑**1.22**) |
|  | SSIM | 0.799 | 0.827(↑**0.028**) | 0.801 | 0.827(↑**0.026**) | 0.801 | 0.825(↑**0.024**) | 0.806 | 0.832(↑**0.026**) |
|  | Time | 0.61 | 0.39(↓**0.22**) | 0.69 | 0.48(↓**0.21**) | 0.90 | 0.70(↓**0.20**) | 7.18 | 6.83(↓**0.35**) |
| head | PSNR | 30.18 | 30.52(↑**0.34**) | 30.18 | 30.53(↑**0.35**) | 30.39 | 30.60(↑**0.21**) | 30.46 | 30.64(↑**0.18**) |
|  | SSIM | 0.726 | 0.743(↑**0.017**) | 0.726 | 0.744(↑**0.018**) | 0.741 | 0.748(↑**0.007**) | 0.746 | 0.752(↑**0.006**) |
|  | Time | 0.68 | 0.45(↓**0.23**) | 0.78 | 0.54(↓**0.24**) | 1.00 | 0.77(↓**0.23**) | 8.50 | 8.32(↓**0.18**) |
| woman | PSNR | 27.71 | 28.90(↑**1.19**) | 27.73 | 28.93(↑**1.20**) | 27.43 | 28.97(↑**1.54**) | 27.70 | 29.10(↑**1.40**) |
|  | SSIM | 0.882 | 0.901(↑**0.019**) | 0.882 | 0.902(↑**0.020**) | 0.885 | 0.904(↑**0.019**) | 0.890 | 0.904(↑**0.014**) |
|  | Time | 0.71 | 0.47(↓**0.24**) | 0.79 | 0.55(↓**0.24**) | 1.04 | 0.80(↓**0.24**) | 8.53 | 7.98(↓**0.55**) |
| average | PSNR | 28.89 | 29.75(↑**0.86**) | 28.90 | 29.76(↑**0.86**) | 28.92 | 29.91(↑**0.99**) | 29.11 | 29.99(↑**0.88**) |
|  | SSIM | 0.838 | 0.856(↑**0.018**) | 0.838 | 0.857(↑**0.019**) | 0.846 | 0.857(↑**0.011**) | 0.849 | 0.859(↑**0.010**) |
|  | Time | 0.88 | 0.59(↓**0.29**) | 1.02 | 0.73(↓**0.29**) | 1.37 | 1.09(↓**0.28**) | 11.77 | 11.40(↓**0.37**) |

## B. Comparison with state-of-the-art methods

To further demonstrate the effectiveness and efficiency of the proposed MISR framework, seven methods are selected to compare with our work. As bicubic interpolation serves as the simplest SR approach, it is selected as a baseline method. Serving as most cited method in the field of MISR for the performance and simplicity, Farsiu's BTV method [25] is selected besides with its variation Li's LABTV method [31]. As the popularity of learning-based approaches increases, we also select Kato's sparse coding method for MISR (denote as MIScSR) [32] as one state-of-the-art method in this field. In addition, Kim's [34] and Yang's [33] methods considered as state-of-the-art SISR methods are also introduced in the comparison.

For fair comparisons, the source codes of Kim's and

Yang's methods released in the authors' homepages are directly implemented in our experiments. As no available codes for other methods, we implement them according to the instructions in [25][31][32], and the performance may differ from the original. Note that Kato's MIScSR method is only utilized for comparison with upscaling factor 3 for instructions in [32] only presented implementation details and parameter settings under this condition. Extensive experiments are conducted on Set14 dataset, and the reconstruction results are exhibited in **Figs. 9-13** with quantitative results presented in **Table. III**.

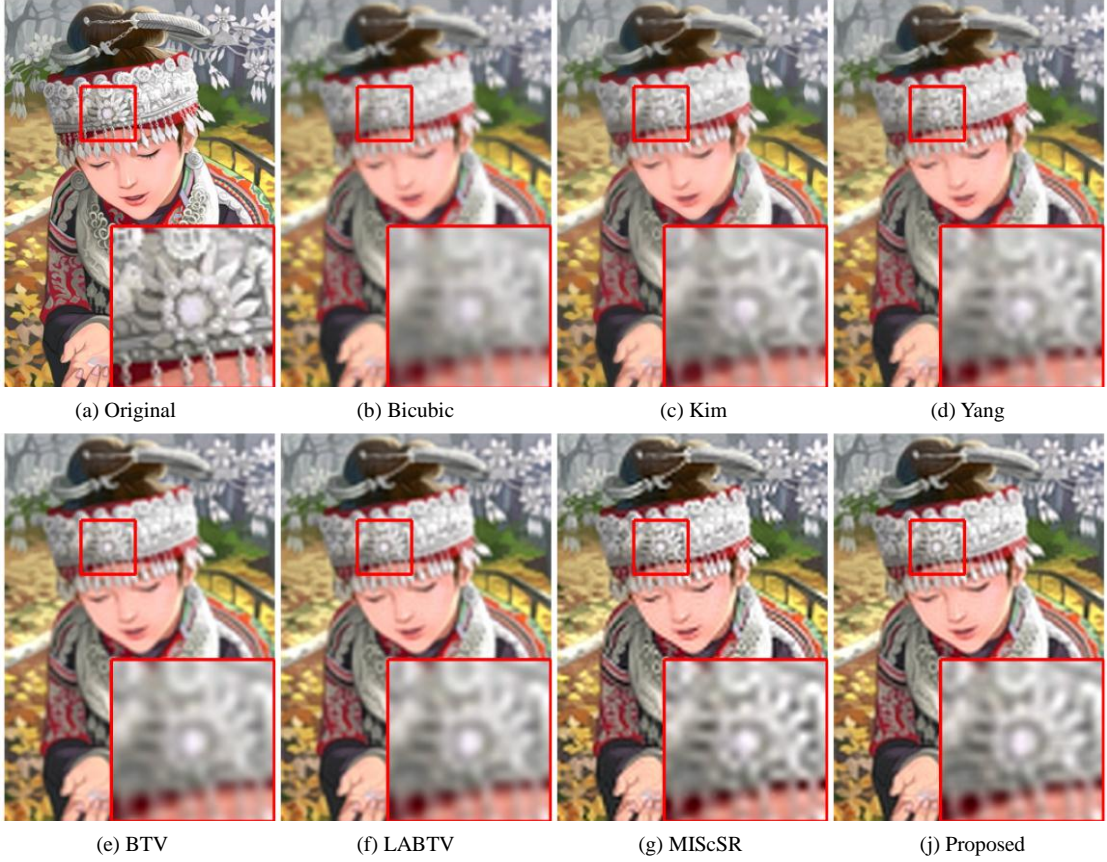

Fig. 10. Reconstruction results for image *Commic* by ranged methods.

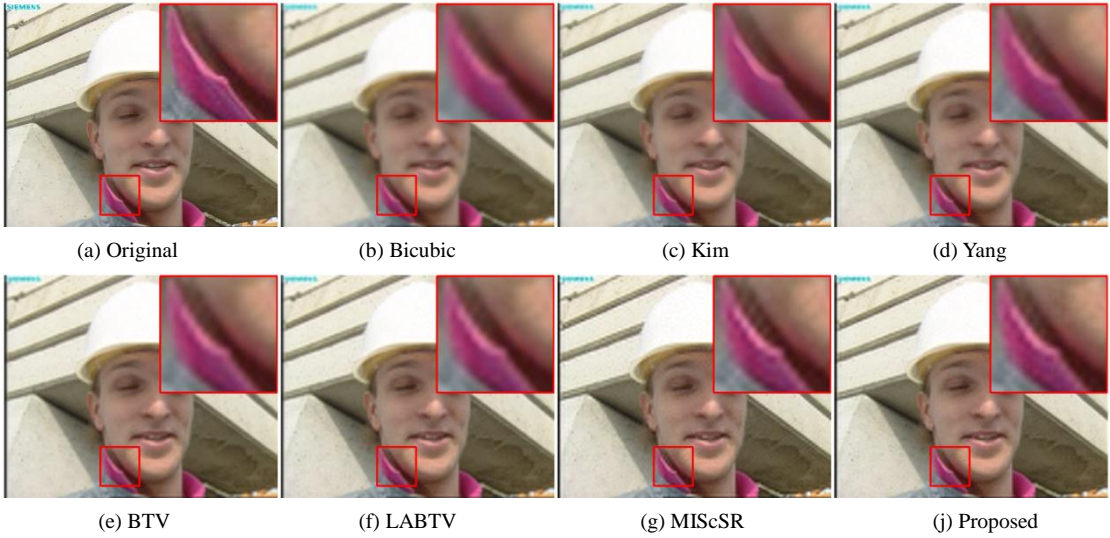

Fig. 11. Reconstruction results for image *Foreman* by ranged methods.

In our scenario, for bicubic method and SISR methods, only the target LR image (first LR image) is utilized for reconstruction, and for MISR methods, same registration procedure is adopted. Note that the blurring kernel is supposed to be given for all SR approaches for its derivation is not the focus of this paper. Other detailed parameter settings for our MISR framework are summarized in **Table. II**.

TABLE II
PARAMETER SETTINGS FOR THE PROPOSED UPSCALING TECHNIQUE

| Parameters | Values |
|---|---|
| $\epsilon$: tolerance threshold | $10^{-5}$ |
| $N_{Max}$: maximum iteration | 30 |
| $\lambda$: regularization parameter in (17) | 0.1 |
| $\eta$: step size in (17) | 0.1 |

From the perspective of visual quality, for SISR methods, Kim's method serving as the superior approach, has already recovered the major structures of the scene, however it tends to oversmooth fine details. For MISR methods, blurring artifacts in BTV and LABTV methods are commonly noticeable, especially within edge and texture regions. Although the sparse representation alleviates the blurring effect for MIScSR, some ragged edges are still visible. By comparison, the proposed MISR approach produces sharper and clearer images with fine details and fewer artifacts.

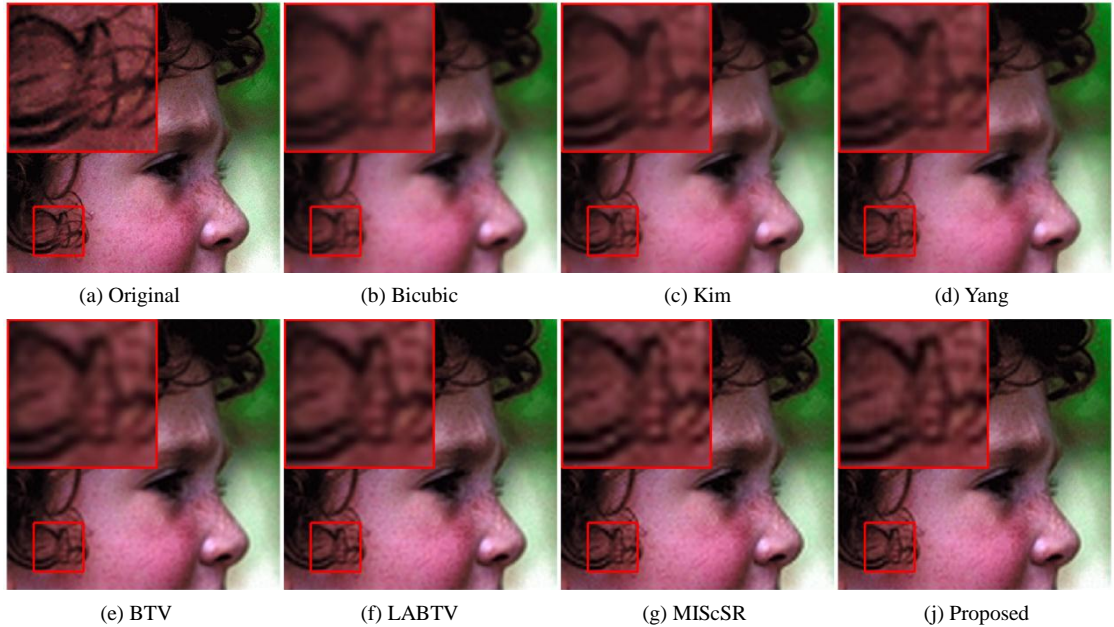

Fig. 12. Reconstruction results for image *Girl* by ranged methods.

From the quantitative results exhibited in **Table. III** we can further see that our approach outperforms other state-of-the-art methods in all the images of Set14 with respect to reconstruction performance and efficiency. Compared with Kim's method, serving as the superior SISR method, the PSNR value of our approach is improved by 0.62 (dB) in average with processing efficiency 10 times faster. Compared with MIScSR method, known as the state-of-the-art MISR method, our approach improves the PSNR value by 0.49 (dB) and runs nearly 8 times faster. Comparing our approach with BTV method, the superiority of the proposed end-to-end upscaling technique can be further extensively validated by the higher PSNR values with shorter running time. Leave out the bicubic method, we can see our approach performs as the most effective and efficient one among comparing methods with practical applications.

TABLE III
COMPARISON OF PSNR AND RUNNING TIME FOR RANGE METHODS. FOR MIMR METHODS, THE MEAN PERFROMANCE OF 50 EXPERIMENTS IS PRESENTED WITH STANDARD DERIVATION SHOWN IN BRACKETS. THE BEST RESULTS ARE SHOWN IN RED BOLD.

| Set14 | Bicubic | | Kim | | Yang | | BTV | | LABTV | | MIScSR | | Proposed | |
|---|---|---|---|---|---|---|---|---|---|---|---|---|---|---|
| | PSNR | Time | PSNR | Time | PSNR | Time | PSNR | Time | PSNR | Time | PSNR | Time | PSNR | Time |
| baboon | 20.81 | 0.04 | 21.19 | 40.16 | 21.16 | 673.24 | 21.24(±0.01) | 3.76 | 21.33(±0.02) | 33.77 | 21.33(±0.00) | 27.95 | **21.64(±0.03)** | **3.01** |
| barbara | 24.13 | 0.04 | 24.76 | 47.94 | 24.75 | 1103.15 | 24.82(±0.02) | 6.59 | 24.95(±0.02) | 61.41 | 25.13(±0.00) | 48.36 | **25.37(±0.04)** | **5.46** |
| bridge | 23.74 | 0.03 | 24.55 | 57.20 | 24.53 | 980.27 | 24.75(±0.02) | 5.47 | 24.92(±0.04) | 57.74 | 24.87(±0.00) | 41.48 | **25.61(±0.06)** | **4.53** |
| coastguard | 24.68 | 0.01 | 25.43 | 18.95 | 25.44 | 319.60 | 25.54(±0.02) | 2.23 | 25.73(±0.04) | 22.66 | 25.60(±0.00) | 15.69 | **26.40(±0.06)** | **1.58** |
| comic | 20.85 | 0.01 | 21.93 | 22.68 | 21.77 | 288.58 | 21.69(±0.01) | 1.94 | 21.86(±0.02) | 19.50 | 22.14(±0.00) | 14.31 | **22.59(±0.04)** | **1.54** |
| face | 29.44 | 0.01 | 30.11 | 10.25 | 29.99 | 275.17 | 30.38(±0.03) | 1.84 | 30.47(±0.04) | 17.23 | 30.22(±0.00) | 12.13 | **30.64(±0.06)** | **1.36** |
| flowers | 24.40 | 0.02 | 25.55 | 32.32 | 25.41 | 424.51 | 25.42(±0.01) | 3.10 | 25.59(±0.02) | 30.09 | 25.73(±0.00) | 24.05 | **26.28(±0.05)** | **2.49** |
| foreman | 28.38 | 0.01 | 30.59 | 12.90 | 30.06 | 253.73 | 29.97(±0.03) | 1.63 | 30.21(±0.05) | 15.69 | 30.43(±0.00) | 11.89 | **31.04(±0.08)** | **1.25** |
| lenna | 28.93 | 0.02 | 30.30 | 18.49 | 29.99 | 412.01 | 30.18(±0.03) | 3.15 | 30.35(±0.04) | 30.35 | 30.32(±0.00) | 21.99 | **30.86(±0.07)** | **2.55** |
| man | 24.93 | 0.02 | 26.13 | 30.87 | 25.92 | 414.53 | 26.00(±0.02) | 2.90 | 26.17(±0.03) | 27.00 | 26.25(±0.00) | 18.70 | **26.88(±0.05)** | **2.37** |
| monarch | 26.81 | 0.03 | 29.02 | 39.32 | 28.39 | 624.27 | 28.13(±0.02) | 4.74 | 28.34(±0.03) | 41.31 | 28.80(±0.00) | 28.23 | **29.30(±0.08)** | **4.61** |
| pepper | 29.18 | 0.02 | 30.30 | 19.08 | 30.10 | 569.97 | 30.19(±0.03) | 3.39 | 30.28(±0.04) | 27.05 | 30.10(±0.00) | 18.74 | **30.54(±0.05)** | **3.02** |
| ppt | 21.31 | 0.02 | 22.91 | 45.44 | 22.52 | 579.91 | 22.10(±0.01) | 4.90 | 22.24(±0.02) | 36.04 | 23.07(±0.00) | 24.78 | **23.11(±0.07)** | **4.24** |
| zebra | 23.77 | 0.02 | 26.05 | 45.29 | 25.58 | 506.18 | 25.41(±0.02) | 2.61 | 25.70(±0.04) | 23.98 | 26.59(±0.00) | 16.43 | **27.21(±0.08)** | **2.73** |
| average | 25.10 | 0.02 | 26.34 | 31.49 | 26.11 | 530.37 | 26.13(±0.02) | 3.45 | 26.30(±0.03) | 31.70 | 26.47(±0.00) | 23.20 | **26.96(±0.06)** | **2.91** |

To further demonstrate the effectiveness of the proposed MISR approach, experiments are conducted on Set5 dataset concerning ranged upscaling factors and noise intensities with results presented in **Table. IV** and **Table. V**.

As shown in **Table. IV**, with upscaling factor increased, the effects of multiple observations are gradually erased due to the growing ill-posedness, leading to the performances of MISR approaches deteriorating severely and even inferior to SISR approaches in several conditions. Although our approach also undergoes same deterioration, it still performs superiorly with highest PSNR values in most conditions.

From **Table. V** we can further see that our approach performs strong robustness and tolerance to noises. As all the SR methods are sensitive to noises and deteriorate with noise intensity increased, the proposed approach still outperforms in average. Even under the condition of noise intensity 3, the proposed approach performs +0.46 dB and +0.42 dB compared with Kim's method and LABTV method respectively.

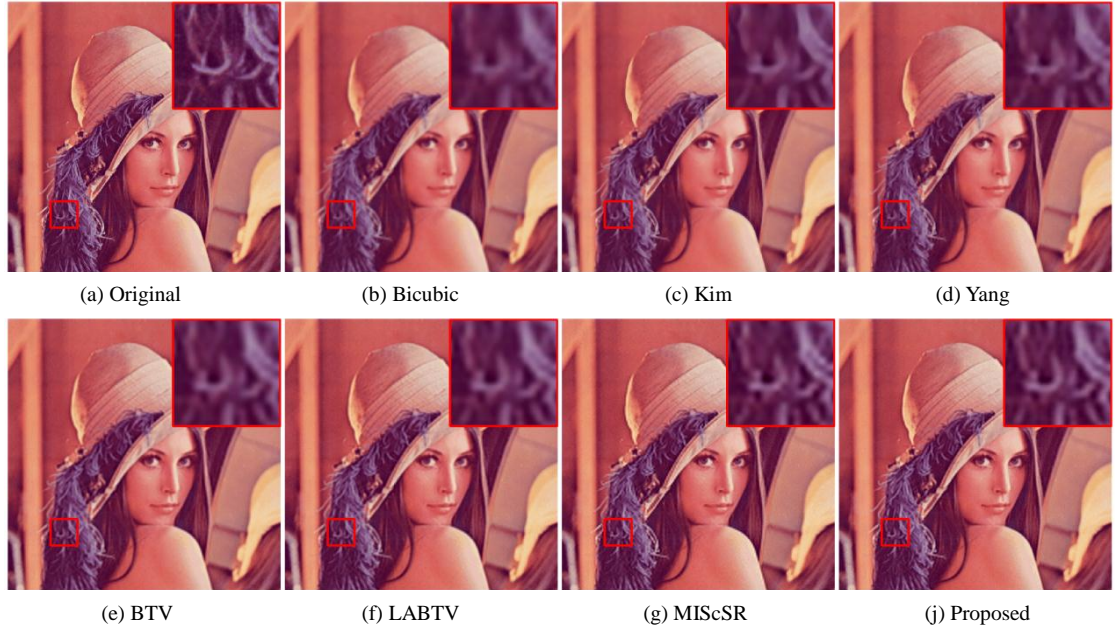

Fig. 13. Reconstruction results for image *Lenna* by ranged methods.
(a) Original (b) Bicubic (c) Kim (d) Yang (e) BTV (f) LABTV (g) MIScSR (j) Proposed

TABLE IV
MAGNIFICATION ×2, ×3 AND ×4 PERFORMANCE IN TERMS OF PSNR ON SET5 DATASET. FOR MIMR METHODS, THE MEAN PERFROMANCE OF 20 EXPERIMENTS IS PRESENTED WITH STANDARD DERIVATION SHOWN IN BRACKETS. THE BEST RESULTS ARE SHOWN IN RED BOLD.

| Set5 | Upscaling factor | Bicubic | Kim | Yang | BTV | LABTV | MIScSR | Proposed |
|---|---|---|---|---|---|---|---|---|
| baby | | 32.53 | 33.52 | 33.30 | 34.74(±0.03) | 35.08(±0.05) | N.A. | **35.37(±0.05)** |
| bird | | 30.71 | 32.00 | 31.68 | 32.89(±0.02) | 33.26(±0.03) | N.A. | **34.20(±0.06)** |
| butterfly | 2 | 22.98 | 24.61 | 24.20 | 24.41(±0.01) | 24.93(±0.02) | N.A. | **26.77(±0.04)** |
| head | | 30.25 | 30.69 | 30.64 | 31.20(±0.03) | 31.37(±0.03) | N.A. | **31.39(±0.03)** |
| woman | | 27.43 | 28.81 | 28.51 | 29.26(±0.02) | 29.79(±0.04) | N.A. | **31.51(±0.07)** |
| average | | 28.78 | 29.93 | 29.67 | 30.50(±0.02) | 30.89(±0.03) | N.A. | **31.85(±0.05)** |
| baby | | 31.07 | 32.39 | 32.23 | 33.13(±0.05) | 33.31(±0.07) | 32.86(±0.00) | **33.83(±0.08)** |
| bird | | 28.94 | 30.58 | 30.22 | 30.81(±0.03) | 31.00(±0.05) | 30.92(±0.00) | **31.87(±0.07)** |
| butterfly | 3 | 21.53 | 24.12 | 23.31 | 22.77(±0.01) | 23.00(±0.02) | 23.57(±0.00) | **24.13(±0.07)** |
| head | | 29.44 | 30.11 | 29.99 | 30.38(±0.03) | 30.47(±0.04) | 30.22(±0.00) | **30.64(±0.06)** |
| woman | | 25.85 | 27.87 | 27.47 | 27.42(±0.03) | 27.67(±0.05) | 28.22(±0.00) | **28.87(±0.13)** |
| average | | 27.38 | 29.02 | 28.65 | 28.91(±0.03) | 29.10(±0.05) | 29.17(±0.00) | **29.87(±0.08)** |
| baby | | 29.63 | 30.08 | 30.77 | 31.60(±0.11) | 31.50(±0.09) | N.A. | **31.85(±0.12)** |
| bird | | 27.31 | 28.76 | 28.49 | 28.88(±0.08) | 28.97(±0.10) | N.A. | **29.45(±0.15)** |
| butterfly | 4 | 20.21 | **22.54** | 21.78 | 21.30(±0.03) | 21.41(±0.04) | N.A. | 22.04(±0.11) |
| head | | 28.75 | 29.39 | 29.21 | 29.61(±0.06) | 29.65(±0.07) | N.A. | **29.66(±0.09)** |
| woman | | 24.48 | 26.37 | 25.86 | 25.81(±0.04) | 25.94(±0.05) | N.A. | **26.66(±0.11)** |
| average | | 26.08 | 27.43 | 27.22 | 27.42(±0.06) | 27.51(±0.07) | N.A. | **27.93(±0.12)** |

TABLE V
NOISE INTENSITY 1, 2 AND 3 PERFORMANCE IN TERMS OF PSNR ON SET5 DATASET. FOR MIMR METHODS, THE MEAN PERFROMANCE OF 50 EXPERIMENTS IS PRESENTED WITH STANDARD DERIVATION SHOWN IN BRACKETS. THE BEST RESULTS ARE SHOWN IN RED BOLD.

| Set5 | Noise intensity | Bicubic | Kim | Yang | BTV | LABTV | MIScSR | Proposed |
|---|---|---|---|---|---|---|---|---|
| baby | | 31.07 | 32.39 | 32.23 | 33.13(±0.05) | 33.31(±0.07) | 32.86(±0.00) | **33.83(±0.08)** |
| bird | | 28.94 | 30.58 | 30.22 | 30.81(±0.03) | 31.00(±0.05) | 30.92(±0.00) | **31.87(±0.07)** |
| butterfly | 1 | 21.53 | 24.12 | 23.31 | 22.77(±0.01) | 23.00(±0.02) | 23.57(±0.00) | **24.13(±0.07)** |
| head | | 29.49 | 30.15 | 30.04 | 30.41(±0.04) | 30.51(±0.04) | 30.27(±0.00) | **30.66(±0.05)** |
| woman | | 25.85 | 27.87 | 27.47 | 27.42(±0.03) | 27.67(±0.05) | 28.22(±0.00) | **28.87(±0.13)** |
| average | | 27.38 | 29.02 | 28.65 | 28.91(±0.03) | 29.10(±0.05) | 29.17(±0.00) | **29.87(±0.08)** |
| baby | | 30.89 | 32.09 | 31.87 | 32.74(±0.04) | 32.89(±0.05) | 32.17(±0.01) | **33.07(±0.06)** |
| bird | | 28.84 | 30.37 | 30.00 | 30.56(±0.04) | 30.73(±0.05) | 30.48(±0.01) | **31.33(±0.08)** |
| butterfly | 2 | 21.50 | **24.07** | 23.27 | 22.72(±0.02) | 22.94(±0.03) | 23.48(±0.00) | 23.93(±0.09) |
| head | | 29.37 | 29.99 | 29.80 | 30.17(±0.03) | 30.25(±0.04) | 29.85(±0.01) | **30.27(±0.04)** |
| woman | | 25.80 | 27.74 | 27.33 | 27.29(±0.03) | 27.53(±0.04) | 27.97(±0.01) | **28.56(±0.09)** |
| average | | 27.28 | 28.85 | 28.45 | 28.70(±0.03) | 28.87(±0.04) | 28.79(±0.01) | **29.43(±0.07)** |
| baby | | 30.63 | 31.61 | 31.38 | 32.24(±0.03) | 32.33(±0.06) | 31.24(±0.03) | **32.36(±0.04)** |
| bird | | 28.69 | 30.04 | 29.68 | 30.25(±0.03) | 30.41(±0.04) | 29.84(±0.02) | **30.82(±0.06)** |
| butterfly | 3 | 21.47 | **23.96** | 23.18 | 22.66(±0.02) | 22.87(±0.03) | 23.34(±0.01) | 23.78(±0.09) |
| head | | 29.18 | 29.64 | 29.49 | 29.90(±0.02) | **29.96(±0.03)** | 29.27(±0.02) | 29.89(±0.03) |
| woman | | 25.71 | 27.52 | 27.12 | 27.14(±0.02) | 27.36(±0.03) | 27.58(±0.01) | **28.24(±0.07)** |
| average | | 27.14 | 28.55 | 28.17 | 28.44(±0.03) | 28.59(±0.04) | 28.25(±0.01) | **29.01(±0.06)** |

## V. CONCLUSIONS

In this paper, we propose an end-to-end fast upscaling technique to replace the interpolation operator for reconstruction-based MISR approaches. As we unfold the degradation model from the perspective of image, we find the influence ranges and underlying contributions of periodic sub-locations vary periodically, which inspires us to design upscaling filters for periodic sub-locations respectively and utilize a shuffling operator to realize effective fusion operation. Equipped with our upscaling technique, remarkable improvements are realized with respect to reconstruction performance and efficiency for reconstruction-based methods, besides the cooperation of our technique and BTV regularization outperforms other state-of-the-art methods demonstrated by extensive experiments.